\documentclass[]{IEEEtran}
\usepackage{cite}
\usepackage{amsmath,amssymb,amsfonts}
\usepackage{algorithmic}
\usepackage{graphicx}
\usepackage{textcomp}
\usepackage{xcolor}
\usepackage{graphicx}
\usepackage{graphics}
\usepackage{multirow}
\usepackage{times}
\usepackage{adjustbox}
\usepackage{amsmath}
\usepackage{amssymb}
\usepackage{array}
\usepackage{color}
\usepackage{subcaption}
\usepackage{mathtools}
\usepackage{url}
\usepackage{bm}
\usepackage{breqn}
\usepackage{chronosys}
\usepackage{xcolor}
\usepackage{multirow}
\usepackage[linesnumbered,ruled,vlined]{algorithm2e}
\usepackage{longtable}
\usepackage{tabu}
\usepackage{lipsum}
\usepackage{tabularx,booktabs}
\usepackage{multirow}

\usepackage{pgfplots}

\pgfplotsset{compat=1.18}
\usepackage{tikz}
\usetikzlibrary{shapes.geometric, arrows, positioning}

\tikzstyle{process} = [rectangle, minimum width=2cm, minimum height=1cm, text centered, draw=black]
\tikzstyle{arrow} = [thick,->,>=stealth]

\begin{document}

\title{Skin Cancer Classification: Hybrid CNN-Transformer Models with KAN-Based Fusion}

\author{\IEEEauthorblockN{Shubhi Agarwal}
\IEEEauthorblockA{\textit{Mehta Family School of Data Science } \\
\textit{and Artificial Intelligence}\\
\textit{Indian Institute of Technology Guwahati}\\
 Assam, India \\
a.shubhi@iitg.ac.in}
\and
\IEEEauthorblockN{Amulya Kumar Mahto}
\IEEEauthorblockA{\textit{Mehta Family School of Data Science } \\
\textit{and Artificial Intelligence}\\
\textit{Indian Institute of Technology Guwahati}\\
Assam, India \\
akmahto@iitg.ac.in}
\and
}

\maketitle

\begin{abstract}
Skin cancer classification is a crucial task in medical image analysis, where precise differentiation between malignant and non-malignant lesions is essential for early diagnosis and treatment. In this study, we explore Sequential and Parallel Hybrid CNN-Transformer models with Convolutional Kolmogorov-Arnold Network (CKAN). Our approach integrates transfer learning and extensive data augmentation, where CNNs extract local spatial features, Transformers model global dependencies, and CKAN facilitates nonlinear feature fusion for improved representation learning.  
To assess generalization, we evaluate our models on multiple benchmark datasets (HAM10000, BCN20000 and PAD-UFES) under varying data distributions and class imbalances. Experimental results demonstrate that hybrid CNN-Transformer architectures effectively capture both spatial and contextual features, leading to improved classification performance. Additionally, the integration of CKAN enhances feature fusion through learnable activation functions, yielding more discriminative representations. Our proposed approach achieves competitive performance in skin cancer classification, demonstrating 92.81\% accuracy and 92.47\% F1-score on the HAM10000 dataset, 97.83\% accuracy and 97.83\% F1-score on the PAD-UFES dataset, and 91.17\% accuracy with 91.79\% F1-score on the BCN20000 dataset highlighting the effectiveness and generalizability of our model across diverse datasets.
This study highlights the significance of feature representation and model design in advancing robust and accurate medical image classification.

\end{abstract}


\section{Introduction}

Cancer remains one of the most pressing global health challenges, with skin cancer being among the most prevalent and rapidly increasing types. Malignant skin cancers, such as melanoma, pose a significant risk due to their aggressive nature and potential for metastasis. Early detection plays a crucial role in improving patient survival rates, and dermoscopic imaging serves as a primary diagnostic tool. However, differentiating malignant from non-malignant lesions remains challenging due to high intra-class variations, inter-class similarities, and dataset biases.

Deep learning, particularly \textbf{Convolutional Neural Networks (CNNs)}, has achieved significant success in medical image classification. While CNNs excel at extracting \textbf{local spatial features}, they struggle to capture \textbf{long-range dependencies}, limiting their ability to analyze complex lesion structures. The inability to model global contextual relationships makes it challenging to distinguish visually similar but clinically different lesions.  
Medical image datasets are often limited in size, leading to overfitting and reduced generalization. \textbf{Transfer learning} has emerged as an effective solution by allowing pre-trained models to adapt to new datasets, improving performance. However, despite these advantages, CNN-based methods still face difficulties in learning robust representations that generalize well across diverse datasets with varying distributions.

To address these limitations, we propose a \textbf{hybrid CNN-Transformer model}, leveraging CNNs for \textbf{hierarchical spatial feature extraction} and Transformers for \textbf{long-range dependency modeling} via self-attention.By combining these architectures, our model enhances feature learning, leading to improved classification performance. Additionally, we introduce the \textbf{Convolutional Kolmogorov-Arnold Network (CKAN)}, which employs \textbf{learnable activation functions} to refine feature fusion and enhance interpretability.

Kolmogorov-Arnold Networks (KANs) differ from conventional deep learning models by replacing fixed activation functions with learnable transformations at the network's edges. 
% This flexibility allows CKAN to dynamically adjust feature representations, improving the adaptability of feature fusion.
In our framework, \textbf{CNNs extract local features, Transformers model global relationships, and CKAN refines feature fusion}, ensuring robust classification. This hierarchical decomposition enhances both \textbf{accuracy and generalization}, making the model more resilient to variations in lesion appearance.

\subsection{Our Contributions}
The key contributions of this study are as follows:
\begin{itemize}
    \item We propose a \textbf{hybrid CNN-Transformer architecture} for skin cancer classification, combining CNNs for local feature extraction with Transformers for global dependency modeling.
    \item We introduce \textbf{CKAN-based feature fusion}, leveraging learnable activation functions to enhance flexibility and feature representation.
    \item We conduct \textbf{comprehensive evaluations} across multiple datasets (HAM10000, BCN20000, ISIC-2020, PAD-UFES) to assess the generalizability of our approach.

    \item We compare \textbf{CNN-only and the hybrid models} to analyze the trade-off between feature extraction techniques and dataset variations.
    \item We provide a detailed analysis of \textbf{the impact of different architectures} on classification performance, highlighting their strengths and limitations in real-world scenarios.
\end{itemize}

\section{Related Works}

Deep learning has significantly advanced medical image analysis, particularly in the classification of skin cancers. CNN-based models have been widely used for feature extraction and classification. The study in \cite{Faghihi2024Diagnosis} applied transfer learning with VGG16 and VGG19, achieving high precision for lesion classification. However, these methods are constrained by their reliance on single datasets, making it challenging to assess their generalizability across diverse real-world data distributions.

Recent works have explored Transformer-based models to address CNNs' limitations in capturing long-range dependencies, Vision Transformers (ViTs) have been explored as an alternative. The review in \cite{Azad2024Advances} highlighted their advantages in learning spatial correlations and dependencies, making them effective for segmentation, detection, and classification tasks. However, ViTs often require large datasets, which can be a challenge in medical imaging. The ViT-CoMer model \cite{Xia_2024_CVPR} introduced a CNN-Transformer fusion mechanism that enhances multiscale feature interaction and bidirectional information exchange, providing a potential improvement in feature representation. Despite these advantages, the direct application of ViT-CoMer to the classification of skin cancer remains an area of further investigation.

Another approach to improving classification performance has focused on enhancing CNN architectures. The work in \cite{ALI2021100036}
proposed a deep CNN model with an extensive preprocessing pipeline, including noise filtering, normalization, and data enhancement, achieving competitive accuracy on HAM10000. However, pre-processing alone may not be sufficient to handle variations across datasets. Meanwhile, methods based on the region of interest (ROI), such as \cite{Ashraf2020}, have attempted to improve classification precision by extracting melanoma-specific regions from images before training CNN models. Although this approach has shown effectiveness, its dependency on accurate segmentation techniques introduces potential bias when applied to varying lesion characteristics and imaging conditions.

Beyond CNNs and Transformers, alternative architectures such as Knowledge-Aware Networks (KANs) have been proposed to enhance feature learning. Unlike traditional CNNs, which use fixed activation functions, KANs employ learnable activation functions at the edges of neural networks, allowing for more flexible feature transformations \cite{Liu2024}. Although KANs have shown promise in various domains, their direct integration into medical image classification remains relatively unexplored. Combining KAN-based feature fusion with existing CNN-Transformer hybrids could provide a more structured approach to improving classification accuracy and interpretability.

Hybrid approaches that integrate CNNs and Transformers have been explored to combine local and global feature extraction capabilities. However, many existing models rely on direct concatenation or simple fusion mechanisms, limiting their effectiveness. The ViTO model \cite{Ovadia2023} applied Transformer-based methods to inverse medical imaging problems, showing strong results for reconstruction tasks. Although not specifically designed for classification, such approaches highlight the potential of structured hybrid models in medical imaging.

Prior studies primarily focus on either CNN-based or Transformer-based architectures, with fewer works exploring structured hybrid models. Publicly available datasets such as HAM10000\cite{tschandl2018ham10000}, BCN20000\cite{combalia2019bcn20000}, ISIC-2020\cite{isic2020}, and PAD-UFES\cite{pacheco2020padufes} have significantly contributed to advancing skin cancer classification research. Although these datasets provide diverse types of lesion and imaging conditions, challenges such as class imbalance and domain shifts persist.

% \begin{figure*}[!ht]
% \centering
% \includegraphics[width=16cm]{myd1.png}
% \caption{A brief of our framework. We have source and target images across four categories, which are plotted as coloured shapes. \textbf{(A)} A common plot indicating their positions. \textbf{(B)} Final plot after adaptation by \pn{}. \textbf{(C)} Test image and train labels are fed to the model for prediction.}
% \label{myfigure}
% \end{figure*}

\section{Methodology}
The proposed approach follows a structured pipeline for classifying skin lesion images into malignant and non-malignant categories. The process begins with data collection from multiple benchmark datasets, ensuring a diverse set of images. Next, preprocessing steps such as resizing and normalization are applied to standardize the data, followed by augmentation techniques like flipping, rotation, and color jitter to improve generalization.

For classification, we evaluate multiple model architectures, including traditional CNNs (ResNet18, ResNet50, AlexNet, MobileNet, DenseNet, EfficientNet-B0) and hybrid models integrating Transformers and KAN layers for enhanced feature fusion. The models undergo a training and validation phase, where optimization techniques such as loss weighting and learning rate scheduling are applied. Finally, model evaluation is performed using accuracy, precision, recall, F1-score to assess performance across datasets.

\noindent
\begin{figure}[ht]
    \centering
    \begin{tikzpicture}[node distance=0.5cm]  % Reduce spacing further
        \node (data) [process] {Data Collection};
        \node (preprocess) [process, below=of data] {Data Preprocessing};
        \node (augment) [process, below=of preprocess] {Data Augmentation};
        \node (models) [process, below=of augment] {Model Selection};
        \node (training) [process, below=of models] {Training and Validation};
        \node (evaluation) [process, below=of training] {Model Evaluation};

        \draw [arrow] (data.south) -- (preprocess.north);
        \draw [arrow] (preprocess.south) -- (augment.north);
        \draw [arrow] (augment.south) -- (models.north);
        \draw [arrow] (models.south) -- (training.north);
        \draw [arrow] (training.south) -- (evaluation.north);
    \end{tikzpicture}
    \caption{Flowchart of the Proposed Methodology}
\end{figure}
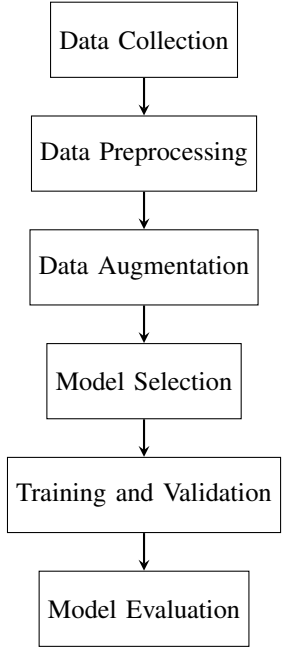

% \begin{center}
%     \begin{tikzpicture}[node distance=1cm]

%         \node (data) [process] {Data Collection};
%         \node (preprocess) [process, right=of data] {Data Preprocessing};
%         \node (augment) [process, right=of preprocess] {Data Augmentation};
%         \node (models) [process, right=of augment] {Model Selection};
%         \node (training) [process, right=of models] {Training and Validation};
%         \node (evaluation) [process, right=of training] {Model Evaluation};

%         \draw [arrow] (data.east) -- (preprocess.west);
%         \draw [arrow] (preprocess.east) -- (augment.west);
%         \draw [arrow] (augment.east) -- (models.west);
%         \draw [arrow] (models.east) -- (training.west);
%         \draw [arrow] (training.east) -- (evaluation.west);

%     \end{tikzpicture}
% \end{center}

\subsection{Dataset}

The primary dataset used for training in this study is HAM10000, a well-established dermatology dataset containing 10,015 dermoscopic images of skin lesions. The dataset includes a diverse range of lesion types, including malignant (melanoma) and benign (nevus, seborrheic keratosis, etc.) cases, making it highly suitable for deep learning-based classification tasks. 

For binary classification, the lesion types were grouped as:

Malignant: Melanoma (mel), Basal Cell Carcinoma (bcc), and Actinic Keratosis (akiec).
Non-Malignant: Nevus (nv), Benign Keratosis-like Lesions (bkl), Vascular Lesions (vasc), and Dermatofibroma (df).

\subsection{Data Processing}
Data processing is a critical step in preparing the dataset for deep learning models, ensuring consistency, efficiency, and optimal model performance. The following steps outline the key preprocessing techniques applied in this study.

\subsubsection{Dataset Splitting}
To ensure a fair evaluation of the model, the dataset is divided into three subsets: \textbf{training, validation, and test sets}. This is achieved using \textbf{stratified sampling}, which maintains the original class distribution across all splits. The process involves:

\begin{itemize}
    \item \textbf{80\% Training Set} – Used to train the deep learning model, allowing it to learn meaningful patterns from the data.
    \item \textbf{10\% Validation Set} – Used to fine-tune hyperparameters, monitor overfitting, and optimize model performance.
    \item \textbf{10\% Test Set} – Held out for the final evaluation to measure the generalization ability of the trained model.
\end{itemize}

\subsubsection{Image Resizing}
The dataset consists of images with varying resolutions. To maintain consistency and compatibility with deep learning models, all images are resized to \textbf{224×224 pixels}. This ensures a uniform input size across the dataset and reduces computational costs without losing significant image details.

\subsubsection{Data Normalization}
Normalization is a crucial preprocessing step that \textbf{scales pixel values} to a standard range, helping to stabilize gradient updates and accelerate convergence during training. The images in this study are normalized using the \textbf{mean and standard deviation of ImageNet}:

\begin{equation}
X_{\text{normalized}} = \frac{X - \mu}{\sigma}
\end{equation}

where:
\begin{itemize}
    \item $X$ is the original pixel value,
    \item $\mu$ is the mean of pixel values across the dataset,
    \item $\sigma$ is the standard deviation of pixel values.
\end{itemize}

For ImageNet normalization, the mean and standard deviation values for the three color channels (RGB) are:

\begin{equation}
\mu = (0.485, 0.456, 0.406), \quad \sigma = (0.229, 0.224, 0.225)
\end{equation}

By centering the data around zero and scaling it, normalization reduces the risk of large weight updates, leading to \textbf{better stability and faster convergence} in deep learning models.

\subsection{Data Augmentation}
Data augmentation is a widely used technique in deep learning to artificially expand the training dataset by introducing transformations to existing images. This helps improve model generalization, reduce overfitting, and make the model more robust to real-world variations in image acquisition. In medical image classification, augmentation is particularly beneficial due to the limited availability of labeled data and the potential for class imbalance .

To improve classification performance and model robustness, we applied multiple augmentation techniques to the training dataset while ensuring that essential lesion characteristics remain preserved. Table~\ref{tab:augmentation_parameters} summarizes the augmentation parameters and their corresponding values used in this study.  

\begin{table}[ht]
\centering
\small
\renewcommand{\arraystretch}{1.5} 
\caption{Data augmentation parameters and values.}
\label{tab:augmentation_parameters}
\begin{tabular}{|m{2cm}|m{2cm}|m{3cm}|}
\hline
\textbf{Data Augmentation Parameter} & \textbf{Parameter Value} & \textbf{Action} \\
\hline
Random Resized Crop & Scale (0.8, 1.0) & Randomly crops a portion of the image and resizes it back to 224×224 pixels, ensuring diverse viewpoints. \\
\hline
Rotation Range & $\pm20^\circ$ & Rotates images randomly within $\pm20^\circ$ to account for different lesion orientations. \\
\hline
Horizontal and Vertical Flip & True & Randomly flips images horizontally to prevent learning orientation-specific biases. \\

\hline
Color Jitter & (Brightness, Contrast, Saturation) & Randomly alters brightness, contrast, and saturation . \\
\hline
Random Grayscale Conversion & - & Converts images to grayscale with a certain probability. \\
\hline
Gaussian Blur & - & Applies Gaussian blur for slight noise simulation. \\
\hline
Normalization & Mean: (0.485, 0.456, 0.406) & Normalizes pixel values to standard mean and variance. \\
\hline
\end{tabular}
\end{table}

\begin{figure}[ht]
\centering
\includegraphics[width=0.5\textwidth]{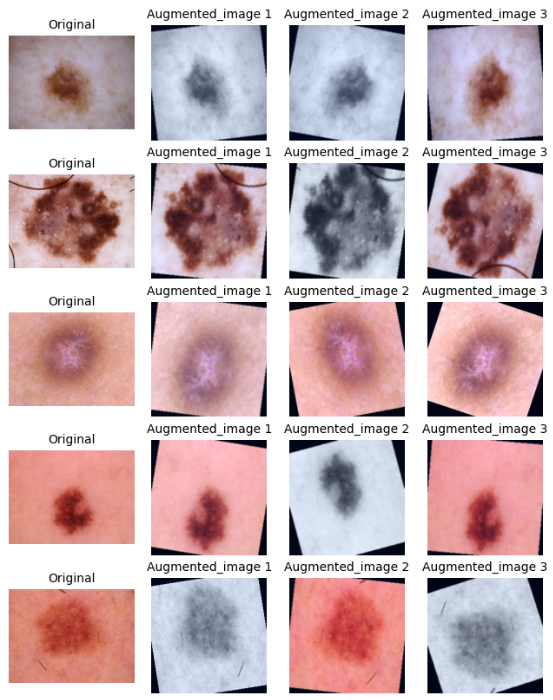} 
\caption{Augmented images.}
\label{fig:yourimage}
\end{figure}

\subsection{Model Description}
In this study, we aim to leverage both convolutional neural networks and transformers to capture short-range and long-range dependencies in skin lesion classification. We explore two approaches: 

\begin{itemize}
    \item \textbf{Sequential Hybrid Model:} Features are first extracted through a Convolutional Neural Network (CNN), then processed by a Transformer for capturing long-range dependencies, and finally classified.
    \item \textbf{Parallel Feature Extraction with Kolmogorov Arnold Network (KAN) Fusion} Features extracted from both CNN and Transformer are concatenated before classification, leveraging both local and global feature representations.
\end{itemize}

\subsubsection{Sequential Model Approach}
In the first approach, the image is passed through a convolutional neural network to extract spatial features. These extracted features are then processed by a transformer encoder to model long-range dependencies. Finally, the output is fed into a fully connected layer for binary classification. 
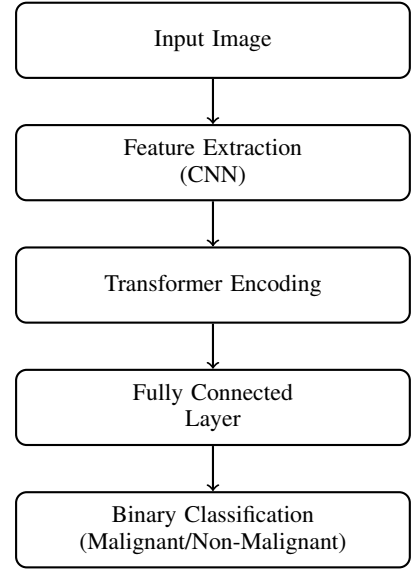
\begin{figure}[ht]
\centering
\begin{tikzpicture}[
    node distance=0.6cm and 1cm,
    every node/.style={draw, minimum height=1cm, minimum width=5cm, align=center, rounded corners, font=\small, text width=5cm},
    every path/.style={draw, ->, thick}
]

% Nodes
\node (input) {Input Image};
\node (cnn) [below=of input] {Feature Extraction \\ (CNN)};
\node (transformer) [below=of cnn] {Transformer Encoding};
\node (fc) [below=of transformer] {Fully Connected \\ Layer};
\node (output) [below=of fc] {Binary Classification \\ (Malignant/Non-Malignant)};

% Arrows
\path (input) -- (cnn);
\path (cnn) -- (transformer);
\path (transformer) -- (fc);
\path (fc) -- (output);

\end{tikzpicture}
\caption{Flowchart for the image classification pipeline.}
\label{fig:Flowchart of model}
\end{figure}

\textit{1.1 Feature Extraction using CNN:}  
The first step in our model involves leveraging \textbf{Convolutional Neural Networks (CNNs)} to extract spatial features from input images. CNNs are particularly effective for medical image analysis due to their ability to capture hierarchical feature representations.  

Key advantages of CNNs in skin lesion classification:  
\begin{itemize}
    \item \textbf{Hierarchical Feature Learning:} Early layers capture low-level patterns such as edges and textures, while deeper layers extract high-level features like lesion shapes and irregularities.
    \item \textbf{Spatial Invariance:} CNNs preserve spatial hierarchies, making them robust to variations in scale, rotation, and illumination.
    \item \textbf{Automated Feature Extraction:} Unlike traditional handcrafted feature methods, CNNs automatically learn the most discriminative patterns from the data.
\end{itemize}

We employ pre-trained CNN architectures such as \textbf{ResNet, MobileNet, EfficientNet, AlexNet, and DenseNet}, which are initialized with weights from large-scale datasets and fine-tuned on our skin lesion dataset to enhance generalization. The extracted features from the CNN are then processed further to incorporate global contextual information.  

\vspace{1pt}
\textit{1.2 Transformer-Based Feature Encoding:}  
After CNN-based feature extraction, the extracted spatial features are processed using a Transformer encoder to capture long-range dependencies and contextual relationships across different regions of the image. Transformers have shown significant advantages in medical image analysis due to their ability to effectively model global dependencies.  

Key benefits of using Transformers:  
\begin{itemize}
    \item \textbf{Self-Attention Mechanism:} Captures global dependencies between different spatial regions of the image, allowing the model to focus on important features irrespective of their position.
    \item \textbf{Positional Encoding:} Since Transformers do not inherently process spatial hierarchies like CNNs, positional encodings are added to retain spatial information.
    \item \textbf{Robust Feature Representation:} The self-attention mechanism enhances feature learning by capturing both local and global contextual relationships.
\end{itemize}

\vspace{1pt}

\textit{1.3 Positional Encoding:}  
Unlike CNNs, which inherently preserve spatial structure through convolutional operations, Transformers process input as a sequence of tokens, losing spatial order. To compensate for this, \textbf{positional encodings} are introduced into the feature representation before applying self-attention. These encodings provide each token with a unique position in the sequence, allowing the Transformer to differentiate between different spatial locations.  

\scriptsize
The positional encoding is defined as follows:  
\begin{equation}
PE_{(pos, 2i)} = \sin\left(\frac{pos}{10000^{2i/d}}\right), \quad
PE_{(pos, 2i+1)} = \cos\left(\frac{pos}{10000^{2i/d}}\right)
\end{equation}
\normalsize

where:
\begin{itemize}
    \item \( pos \) represents the position of the token in the sequence,
    \item \( i \) is the dimension index,
    \item \( d \) is the total feature dimension.
\end{itemize}

By adding these positional embeddings to the extracted CNN features, the model retains spatial relationships while leveraging the Transformer’s powerful self-attention mechanism.

\textit{1.3 Classification Layer:}  
The final step in the sequential model is the classification layer, which consists of:
\begin{itemize}
    \item A \textbf{fully connected (FC) layer} that maps the transformer-processed feature representation into a binary classification output.
    \item A \textbf{sigmoid activation function} applied to produce a probability score indicating the likelihood of malignancy.
    \item A \textbf{binary cross-entropy loss function} used for optimization.
\end{itemize}

The classification layer enables the model to output a final decision regarding whether the skin lesion is malignant or non-malignant.

\vspace{1\baselineskip} 

\subsubsection{Parallel Feature Extraction with KAN Fusion}
In this approach, feature extraction is performed using a \textbf{parallel processing pipeline} where \textbf{CNN} and \textbf{Transformer} modules independently extract representations from the input image. Unlike the sequential model, where the Transformer refines CNN features, this approach \textbf{combines both feature types} before passing them through a \textbf{Kolmogorov-Arnold Network (KAN) fusion layer}. The goal is to leverage \textbf{both local and global feature representations}, ensuring that spatial details captured by CNNs and contextual relationships modeled by Transformers contribute effectively to the classification task.

After feature extraction, the CNN and Transformer outputs are concatenated and passed through the KAN layer, which refines the joint representation before feeding it into the fully connected layer for classification.

\textbf{Key Components of this Approach:}
\begin{itemize}
    \item \textbf{Parallel Feature Learning:} CNN extracts local spatial features while the Transformer processes long-range dependencies simultaneously.
    \item \textbf{KAN-Based Fusion:} The Kolmogorov-Arnold Network learns a nonlinear mapping to refine and integrate the combined feature space.
\end{itemize}

\textbf{Feature Extraction Process:}
\begin{enumerate}
    \item The input image is passed through a CNN , which outputs \textbf{local feature embeddings} \( F_{CNN} \).
    \item Simultaneously, the Transformer encoder processes the input image, generating \textbf{global feature embeddings} \( F_{Trans} \).
    \item The extracted features are concatenated:  
    \begin{equation}
    F_{concat} = [F_{CNN}; F_{Trans}]
    \end{equation}
    \item The concatenated features pass through the \textbf{KAN Layer}, which applies nonlinear transformations to learn complex dependencies:
    \begin{equation}
    F_{KAN} = \sigma(W_2 \cdot \text{ReLU}(W_1 \cdot F_{concat}))
    \end{equation}
    where \( W_1 \) and \( W_2 \) are learnable weight matrices.
    \item The fused representation is passed through a \textbf{fully connected layer}, followed by a \textbf{sigmoid activation} to obtain the final classification.
\end{enumerate}

\begin{figure}[ht]
\centering
\begin{tikzpicture}[
    node distance=0.5cm and 0.01cm, % Adjusted spacing
    smallbox/.style={draw, minimum height=0.8cm, align=center, rounded corners, font=\small, text width=2.5cm},
    largebox/.style={draw, minimum height=0.8cm, align=center, rounded corners, font=\small, text width=3.0cm},
    every path/.style={draw, ->, thick}
]

% Nodes
\node (input) [largebox] {Input Image};
\node (cnn) [smallbox, below left=of input, xshift=-0.1cm] {Feature Extraction \\ (CNN)};
\node (transformer) [smallbox, below right=of input, xshift=0.1cm] {Feature Extraction (Transformer)};
\node (kan) [largebox, below=of input, yshift=-1.8cm] {Feature Fusion \\ (KAN)};
\node (fc) [largebox, below=of kan] {Fully Connected \\ Layer};
\node (output) [largebox, below=of fc] {Binary Classification \\ (Malignant/Non-Malignant)};

% Arrows
\path (input) -- (cnn);
\path (input) -- (transformer);
\path (cnn) -- (kan);
\path (transformer) -- (kan);
\path (kan) -- (fc);
\path (fc) -- (output);

\end{tikzpicture}
\caption{Flowchart for the Parallel Feature Extraction with KAN Fusion model.}
\label{fig:KAN_model}
\end{figure}
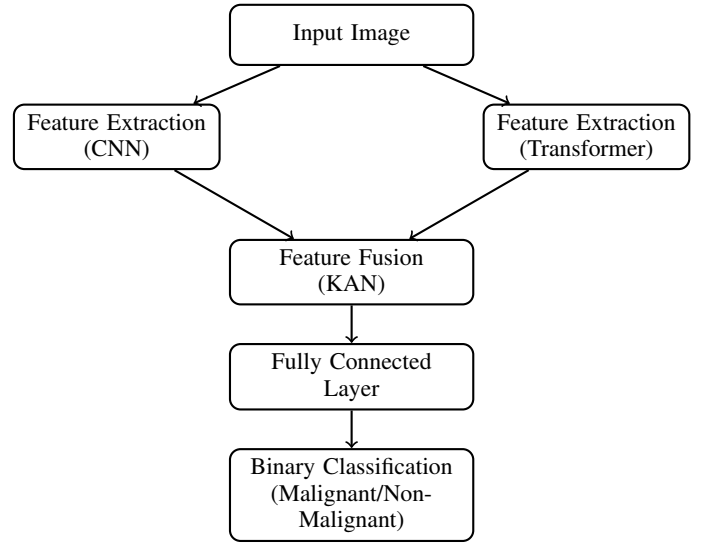

This \textbf{KAN-enhanced fusion model} effectively integrates CNN and Transformer features to improve classification performance by preserving both \textbf{local spatial} and \textbf{global contextual} information. The next section details the evaluation methodology and comparative analysis across different models.

The sequential model allows the transformer to refine CNN-extracted features progressively, whereas the KAN-based approach explicitly combines features, leveraging both local and global information before classification. We compare both models in terms of classification performance across different datasets.

\subsection{Algorithm}
The pseudo code of the algorithm is given as-

\newcommand\mycommfont[1]{\footnotesize\ttfamily\textcolor{blue}{#1}}

\SetCommentSty{mycommfont}
\SetKwInput{KwInput}{Input}                % Set the Input
\SetKwInput{KwOutput}{Output}              % Set the Output
\begin{algorithm}[!ht]
\DontPrintSemicolon
  \KwInput
  {\\
  Training Dataset $D = \{(I_i, y_i)\}_{i=1}^{N}$\\
  Pretrained CNN Backbone $CNN_{\theta}$\\
  Transformer Encoder $T_{\phi}$\\
  Positional Encoding Module $PE$\\
  Fully Connected Layer $FC_{\omega}$\\
  Binary Cross-Entropy Loss $L_{BCE}$\\
  Learning Rate $\eta$, Batch Size $B$, Epochs $E$
  }
  \KwOutput
  {\\
  Trained model parameters $\{\theta, \phi, \omega\}$ 
  }

  \For{$epoch = 1$ \KwTo $E$}
   {
       \For{$batch = 1$ \KwTo $\frac{N}{B}$}
       {
            % // Sample a batch of images and labels
            $I_{batch}, y_{batch} \gets \text{SampleBatch}(D, B)$

            % // Extract features using CNN
            $F_{CNN} \gets CNN_{\theta}(I_{batch})$

            % // Reshape CNN features for Transformer input
            $F_{CNN} \gets \text{Reshape}(F_{CNN})$

            % // Apply Positional Encoding
            $F_{PE} \gets PE(F_{CNN})$

            % // Transformer-Based Feature Encoding
            $F_{T} \gets T_{\phi}(F_{PE})$

            % // Aggregate features (e.g., Mean Pooling)
            $F_{agg} \gets \text{MeanPooling}(F_{T})$

            % // Classification Layer
            $Output \gets FC_{\omega}(F_{agg})$

            % // Compute BCE Loss
            $\mathcal{L} = L_{BCE}(Output, y_{batch})$

            % // Backpropagation
            \text{Compute gradients} $\nabla_{\theta, \phi, \omega} \mathcal{L}$  
            \text{Update parameters:}  
            $\theta, \phi, \omega \gets \theta, \phi, \omega - \eta \nabla_{\theta, \phi, \omega} \mathcal{L}$
       }

       % // Validation Step (if needed)
       $\mathcal{L}_{val} = \text{Evaluate Model on Validation Set}$

       % // Apply Learning Rate Scheduling
       $\eta \gets \text{Adjust Learning Rate (if scheduled)}$
   }

   \KwRet $\{\theta, \phi, \omega\}$ (Trained Model Parameters)

\caption{Pseudocode for Training Hybrid CNN-Transformer Model with Positional Encoding}
\label{alg:Hybrid_Training}
\end{algorithm}

% \SetCommentSty{mycommfont}
% \SetKwInput{KwInput}{Input}                % Set the Input
% \SetKwInput{KwOutput}{Output}              % Set the Output
% \begin{algorithm}[!ht]
% \DontPrintSemicolon
%   \KwInput
%   {\\
%   Input Image $I$\\
%   Pretrained CNN Backbone $CNN_{\theta}$\\
%   Transformer Encoder $T_{\phi}$\\
%   Kolmogorov-Arnold Network (KAN) Layer $KAN_{\psi}$\\
%   Fully Connected Layer $FC_{\omega}$
%   }
%   \KwOutput
%   {\\
%   Binary Classification Output: Malignant or Non-Malignant
%   }

%    % // Extract features using CNN
%    $F_{CNN}$ = $CNN_{\theta}(I)$ 

%    % // Extract features using Transformer
%    $F_{T}$ = $T_{\phi}(I)$  

%    % // Concatenate CNN and Transformer features
%    $F_{concat}$ = $[F_{CNN}, F_{T}]$  

%    % // Apply KAN for feature fusion
%    $F_{KAN}$ = $KAN_{\psi}(F_{concat})$  

%    % // Classification Layer
%    $Output$ = $FC_{\omega}(F_{KAN})$  
%    $\hat{y}$ = $\sigma(Output)$  

%    \KwRet $\hat{y}$ (Predicted class: Malignant or Non-Malignant)

% \caption{Pseudocode for Parallel Feature Extraction with KAN Fusion}
% \label{alg:KAN_Model}
% \end{algorithm}

\SetCommentSty{mycommfont}
\SetKwInput{KwInput}{Input}                % Set the Input
\SetKwInput{KwOutput}{Output}              % Set the Output
\begin{algorithm}[!ht]
\DontPrintSemicolon
  \KwInput
  {\\
  Training Dataset $D = \{(I_i, y_i)\}_{i=1}^{N}$\\
  Pretrained CNN Backbone $CNN_{\theta}$\\
  Transformer Encoder $T_{\phi}$\\
  Kolmogorov-Arnold Network (KAN) Layer $KAN_{\psi}$\\
  Fully Connected Layer $FC_{\omega}$\\
  Binary Cross-Entropy Loss $L_{BCE}$\\
  Learning Rate $\eta$, Batch Size $B$, Epochs $E$
  }
  \KwOutput
  {\\
  Trained model parameters $\{\theta, \phi, \psi, \omega\}$ 
  }

  \For{$epoch = 1$ \KwTo $E$}
   {
       \For{$batch = 1$ \KwTo $\frac{N}{B}$}
       {
            % // Sample a batch of images and labels
            $I_{batch}, y_{batch} \gets \text{SampleBatch}(D, B)$

            % // Extract features using CNN
            $F_{CNN} \gets CNN_{\theta}(I_{batch})$

            % // Extract features using Transformer
            $F_{T} \gets T_{\phi}(I_{batch})$

            % // Concatenate CNN and Transformer features
            $F_{concat} \gets [F_{CNN}, F_{T}]$

            % // Apply KAN for feature fusion
            $F_{KAN} \gets KAN_{\psi}(F_{concat})$

            % // Classification Layer
            $Output \gets FC_{\omega}(F_{KAN})$

            % // Compute BCE Loss
            $\mathcal{L} = L_{BCE}(Output, y_{batch})$

            % // Backpropagation
            \text{Compute gradients} $\nabla_{\theta, \phi, \psi, \omega} \mathcal{L}$  
            \text{Update parameters:}  
            $\theta, \phi, \psi, \omega \gets \theta, \phi, \psi, \omega - \eta \nabla_{\theta, \phi, \psi, \omega} \mathcal{L}$
       }

       % // Validation Step (if needed)
       $\mathcal{L}_{val} = \text{Evaluate Model on Validation Set}$

       % // Apply Learning Rate Scheduling
       $\eta \gets \text{Adjust Learning Rate (if scheduled)}$
   }

   \KwRet $\{\theta, \phi, \psi, \omega\}$ (Trained Model Parameters)

\caption{Pseudocode for Training Parallel Feature Extraction with KAN Fusion}
\label{alg:KAN_Training}
\end{algorithm}

\subsection{Training and Validation}

To ensure optimal performance and generalization, the models undergo a structured training and validation process. The training phase involves feeding the input images into the network, computing loss, and updating the model parameters through backpropagation. The validation phase evaluates model performance on unseen data to monitor generalization and prevent overfitting.

\textbf{Training Procedure:} The model is trained using mini-batch gradient descent with a batch size of 32. The training dataset is used to iteratively optimize model parameters, while validation is performed at the end of each epoch to assess performance stability.

\textbf{Loss Function:} Given the binary classification task (malignant vs. non-malignant), the Binary Cross-Entropy (BCE) loss function is used:
\begin{equation}
    L_{BCE} = -\frac{1}{N} \sum_{i=1}^{N} \left[ y_i \log (\hat{y}_i) + (1 - y_i) \log (1 - \hat{y}_i) \right]
\end{equation}
where $y_i$ represents the true label, $\hat{y}_i$ is the predicted probability, and $N$ is the total number of samples in the batch.

\textbf{Optimizer:} The model is optimized using the Adam optimizer with weight decay regularization to prevent overfitting:
\begin{equation}
    \theta_{t+1} = \theta_t - \eta \cdot \frac{m_t}{\sqrt{v_t} + \epsilon}
\end{equation}
where $\theta_t$ represents the model parameters at time step $t$, $\eta$ is the learning rate, and $m_t$ and $v_t$ are the first and second moment estimates.

\textbf{Learning Rate Scheduling:} A StepLR scheduler is employed, which reduces the learning rate by a factor of 0.5 every 5 epochs to enhance convergence.

\textbf{Class Imbalance Handling:} Given the imbalance in malignant and non-malignant cases, class weights are assigned as:
\begin{equation}
    w_0 = \frac{N}{2 \times N_0}, \quad w_1 = \frac{N}{2 \times N_1}
\end{equation}
where $N_0$ and $N_1$ represent the number of non-malignant and malignant samples, respectively.

\textbf{Validation Process:} The model's performance is evaluated using accuracy, precision, recall, and F1-score. The best model is selected based on the highest validation F1-score.

\subsection{Evaluation Metrics}
To assess the performance of the proposed models, we utilize standard classification metrics, including accuracy, precision, recall, F1-score, and area under the receiver operating characteristic curve (AUC-ROC). Given the inherent class imbalance in the dataset, we compute \textbf{weighted} versions of these metrics to ensure that both classes contribute proportionally to the final evaluation scores.

\textbf{Accuracy:}  
Accuracy measures the overall correctness of the model’s predictions and is defined as:

\begin{equation}
\text{Accuracy} = \frac{TP + TN}{TP + TN + FP + FN}
\end{equation}

where TP, TN, FP, and FN represent true positives, true negatives, false positives, and false negatives, respectively. While accuracy provides a general measure of model performance, it can be misleading in imbalanced datasets, as a model biased toward the majority class may still achieve a high accuracy.

\textbf{Precision (Positive Predictive Value):}  
Precision quantifies the proportion of correctly identified malignant cases among all predicted malignant cases:

\begin{equation}
\text{Precision} = \frac{TP}{TP + FP}
\end{equation}

Higher precision indicates a lower false positive rate, which is critical in medical applications where misclassifying a benign lesion as malignant can lead to unnecessary interventions.

\textbf{Recall (Sensitivity or True Positive Rate):}  
Recall measures the model’s ability to correctly identify malignant cases among all actual malignant cases:

\begin{equation}
\text{Recall} = \frac{TP}{TP + FN}
\end{equation}

A high recall ensures that the model minimizes false negatives, reducing the risk of missing malignant lesions.

\textbf{F1-Score:}  
F1-score provides a harmonic mean between precision and recall, offering a balanced measure when there is an imbalance between the two:

\begin{equation}
\text{F1-Score} = 2 \times \frac{\text{Precision} \times \text{Recall}}{\text{Precision} + \text{Recall}}
\end{equation}

This metric is particularly useful when dealing with imbalanced datasets, as it considers both false positives and false negatives.

\textbf{Weighted Metrics for Class Imbalance:}  
Since the dataset exhibits a class imbalance, we compute weighted versions of precision, recall, and F1-score. These weighted metrics are calculated as:

\begin{equation}
\text{Metric}_{\text{weighted}} = \sum_{i \in \{malignant, non\text{-}malignant\}} w_i \times \text{Metric}_i
\end{equation}

where \(w_i\) represents the proportion of each class in the dataset. This ensures that the evaluation results are not biased toward the majority class and that performance is fairly measured across both classes.

\begin{figure}[ht]
\centering
\includegraphics[width=0.5\textwidth]{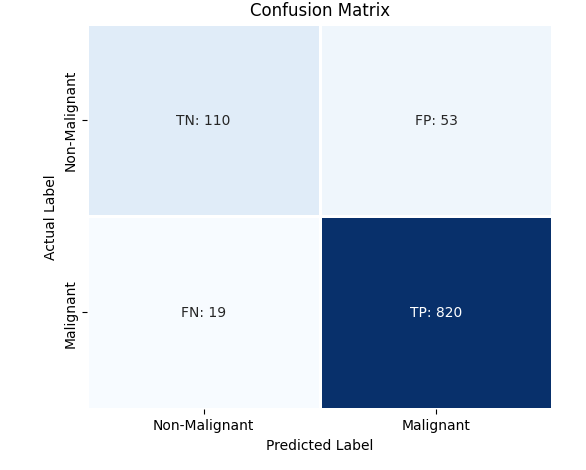} 
\caption{Confusion Matrix of Sequential model using Resnet18.}
\label{fig:confusion matrix}
\end{figure}

\section{Experiment and Analysis}

In this section, we evaluate the performance of our proposed approach on skin lesion classification across multiple datasets. We compare our model with various baseline architectures and analyze its effectiveness in distinguishing malignant and benign lesions. We have trained our models for 10 epochs.

\subsection{Description of Dataset}

We conduct experiments using four publicly available skin lesion datasets: HAM10000,  ISIC-2020, PAD-UFES, and BCN20000 for evaluation to assess the model's robustness across different imaging conditions and class distributions. These datasets introduce variations in image resolution, acquisition methods, and class imbalances, providing a comprehensive evaluation setup.

\begin{itemize}
    \item \textbf{HAM10000:} Contains 10,015 images across seven diagnostic categories, with a focus on melanoma and non-melanoma lesions.
    \item \textbf{ISIC-2020:} A large dataset with 40,000+ images, covering a wide range of skin diseases, including melanoma.
    \item \textbf{PAD-UFES:} A relatively smaller dataset with 2,298 images, including high variability in lesion appearance.
    \item \textbf{BCN20000:} Contains 13,786 dermoscopic images, collected from multiple clinical sources.
\end{itemize}

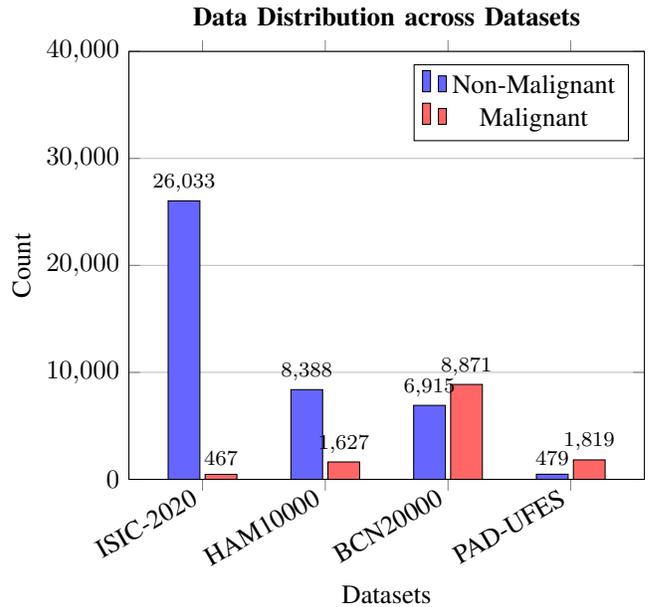
\begin{figure}[ht]
    \centering
    \hspace{-20pt} % Adjust for left shift if needed
    \begin{tikzpicture}
        \begin{axis}[
            title={\textbf{Data Distribution across Datasets}}, % Title of the plot
            ybar,
            symbolic x coords={ISIC-2020, HAM10000, BCN20000, PAD-UFES}, % Ordered by non-malignant cases
            xtick=data,
            ymin=0,
            ymax=40000,
            ylabel={Count},
            xlabel={Datasets}, % Keep x-axis title horizontal
            ylabel style={yshift=-1pt}, 
            xlabel style={yshift=1pt}, % Increased spacing between x-axis labels and x-axis title
            legend pos=north east, 
            bar width=12pt,            
            enlarge x limits=0.20,  % Adjust spacing between bars
            x tick label style={rotate=30, anchor=east}, % Rotate dataset names only
            ymajorgrids=true, % Add horizontal grid lines
            nodes near coords,  % Show values on bars
            every node near coord/.append style={font=\footnotesize, black}, % Adjust label font
            scaled y ticks=false % Prevent scientific notation on y-axis
        ]
            \addplot[fill=blue!60] coordinates {(ISIC-2020, 26033) (HAM10000, 8388) (BCN20000, 6915) (PAD-UFES, 479)};
            \addplot[fill=red!60] coordinates {(ISIC-2020, 467) (HAM10000, 1627) (BCN20000, 8871) (PAD-UFES, 1819)};
            \legend{Non-Malignant, Malignant}
        \end{axis}
    \end{tikzpicture}
    \caption{Data distribution across datasets.} % Figure caption
    \label{fig:data-distribution} % Reference label
\end{figure}

\subsection{Results and Discussion}

The evaluation across multiple datasets highlights how model architecture, dataset size, and class distribution influence classification performance. Our experiments compare CNN-only models, Sequential Hybrid CNN-Transformer models, and Parallel Hybrid CNN-Transformer with CKAN models to analyze their effectiveness in different scenarios.

Among all models, the Sequential Hybrid CNN-Transformer with EfficientNet-B0 achieved the highest accuracy (92.81\%) and F1-score (92.47\%) on HAM10000, demonstrating that integrating Transformers after CNN feature extraction enhances global feature representation. The self-attention mechanism allows the model to capture long-range dependencies, refining CNN-extracted spatial features, which is crucial for distinguishing subtle lesion differences. Notably, Vision Transformer (ViT) also performed well among base models, highlighting that Transformers alone can achieve competitive results. However, its performance was still lower than hybrid models, suggesting that CNNs remain essential for local feature extraction in skin cancer classification.

For PAD-UFES, the Parallel Hybrid CNN-Transformer with CKAN using ResNet18 outperformed other models, achieving an accuracy of 97.83\%. This suggests that feature fusion with KAN is beneficial when CNN and Transformer features contain complementary information. KAN layers provide nonlinear feature transformations, improving feature expressiveness, particularly in datasets with limited training samples. DenseNet121 and EfficientNet-B0 consistently performed well, benefiting from dense connectivity and parameter efficiency, whereas AlexNet showed the lowest accuracy, likely due to its shallower architecture and limited feature extraction capability.

Across datasets, Sequential Hybrid models consistently outperformed CNN-only models, showcasing that progressively refining CNN features with Transformer attention mechanisms leads to better feature learning. However, Parallel Hybrid models excelled in datasets with high class imbalance, demonstrating the advantage of explicit feature fusion through KAN when feature representations from CNN and Transformer differ significantly.

Another key observation is that ResNet-based architectures (ResNet18, ResNet50) improved more significantly in hybrid models compared to standalone CNNs. This suggests that ResNet’s hierarchical feature maps complement Transformer-based feature encoding, enhancing feature discrimination. MobileNetV2, which is optimized for efficiency, showed the least improvement in hybrid models, indicating that lightweight CNNs might not provide sufficiently rich feature representations for Transformers to refine effectively.

These findings emphasize that no single model is universally optimal, and the choice of architecture should be dataset-dependent. Hybrid models incorporating both CNN and Transformer components offer better generalization, especially when tailored to dataset size and class imbalance characteristics.

This multi-dataset analysis provides insights into the generalization ability of each approach, highlighting their strengths and limitations in real-world clinical scenarios.

\begin{table}[ht]
    \centering
    \renewcommand{\arraystretch}{1.5}  % Increase row spacing
    \caption{Performance comparison of different models on HAM10000 dataset}
    \begin{tabular}{|c|c|c|c|c|}
        \hline
        \textbf{Model} & \textbf{Precision} & \textbf{Recall} & \textbf{F1 Score} & \textbf{Accuracy} \\
        \hline
        \multicolumn{5}{|c|}{\textbf{Base Models (CNN-only)}} \\
        \hline
        ResNet18 \cite{he2016deep} & 0.8960 & 0.9022 & 0.8968 & 0.9022 \\
        
        EfficientNet-B0 \cite{tan2019efficientnet} & 0.9026 & 0.9078 & 0.9040 & 0.9078 \\

        ResNet50 \cite{he2016deep}  & 0.9021 & 0.9062 & 0.9034 & 0.9062 \\
        
        AlexNet \cite{krizhevsky2012imagenet}   & 0.8926 & 0.8992 & 0.8988 & 0.8992 \\
        DenseNet121 \cite{huang2017densely} & 0.8917 & 0.8972 & 0.8934 & 0.8972 \\
        MobileNetV2 \cite{sandler2018mobilenetv2} & \textcolor{blue}{0.9085} & \textcolor{blue}{0.9132} & \textcolor{blue}{0.9086} & \textcolor{blue}{0.9132} \\

        \hline\hline
        \multicolumn{5}{|c|}{\textbf{Sequential Hybrid CNN-Transformer}} \\
        \hline
        ResNet18 \cite{he2016deep} & \textcolor{blue}{0.9187} & \textcolor{blue}{0.9227} & \textcolor{blue}{0.9197} & \textcolor{blue}{0.9227} \\
        EfficientNet-B0 \cite{tan2019efficientnet} & \textcolor{blue}{\textbf{0.9252}} & \textcolor{blue}{\textbf{0.9281}} & \textcolor{blue}{\textbf{0.9247}} & \textcolor{blue}{\textbf{0.9281}} \\
        ResNet50 \cite{he2016deep} & \textcolor{blue}{0.9107} & \textcolor{blue}{0.9152} & \textcolor{blue}{0.9104} & \textcolor{blue}{0.9152} \\
        
        AlexNet \cite{krizhevsky2012imagenet} & \textcolor{blue}{0.8989} & \textcolor{blue}{0.9042} & \textcolor{blue}{0.9002} & \textcolor{blue}{0.9042} \\
        
        DenseNet121 \cite{huang2017densely} & \textcolor{blue}{0.9030} & \textcolor{blue}{0.9062} & \textcolor{blue}{0.9042} & \textcolor{blue}{0.9062} \\
        MobileNetV2 \cite{sandler2018mobilenetv2} & 0.9016 & 0.9072 & 0.9016 & 0.9072 \\
        \hline\hline
        \multicolumn{5}{|c|}{\textbf{Parallel Hybrid CNN-Transformer with CKAN}} \\
        \hline
        ResNet18 \cite{he2016deep} & 0.8964 & 0.9002 & 0.8978 & 0.9002 \\
        EfficientNet-B0 \cite{tan2019efficientnet} & 0.9025 & 0.9052 & 0.9036 & 0.9052 \\
        ResNet50 \cite{he2016deep} & 0.9062 & 0.9112 & 0.9050 & 0.9112 \\
        AlexNet \cite{krizhevsky2012imagenet} & 0.8864 & 0.8872 & 0.8868 & 0.8872 \\
        DenseNet121 \cite{huang2017densely} & 0.8987 & 0.9005 & 0.8990 & 0.9005 \\
        MobileNetV2 \cite{sandler2018mobilenetv2} & 0.9034 & 0.9082 & 0.9001 & 0.9082 \\
        \hline
    \end{tabular}
\end{table}

\begin{table}[ht]
    \centering
    \renewcommand{\arraystretch}{1.5}  % Increase row spacing
    \caption{Performance comparison of different models on PAD-UFES dataset}
    \begin{tabular}{|c|c|c|c|c|}
        \hline
        \textbf{Model} & \textbf{Precision} & \textbf{Recall} & \textbf{F1 Score} & \textbf{Accuracy} \\
        \hline
        \multicolumn{5}{|c|}{\textbf{Base Models (CNN-only)}} \\
        \hline
        ResNet18 \cite{he2016deep} & 0.9506 & 0.9435 & 0.9452 & 0.9435 \\
        ResNet50 \cite{he2016deep} & \textcolor{blue}{0.9488} & \textcolor{blue}{0.9478} & \textcolor{blue}{0.9482} & \textcolor{blue}{0.9478} \\
        EfficientNet-B0 \cite{tan2019efficientnet} & 0.9332 & 0.9304 & 0.9314 & 0.9304 \\
        AlexNet \cite{krizhevsky2012imagenet} & 0.9332 & 0.9304 & 0.9314 & 0.9304 \\
        DenseNet121 \cite{huang2017densely} & {0.9586} & {0.9565} & {0.9571} & {0.9565} \\
        MobileNetV2 \cite{sandler2018mobilenetv2} & 0.9350 & 0.9304 & 0.9319 & 0.9304 \\
        
        \hline\hline
        \multicolumn{5}{|c|}{\textbf{Sequential Hybrid CNN-Transformer}} \\
        \hline
        ResNet18 \cite{he2016deep} & 0.9568 & 0.9522 & {0.9533} & 0.9522 \\
        ResNet50 \cite{he2016deep} & 0.9366 & 0.9348 & 0.9355 & 0.9348 \\
        EfficientNet-B0 \cite{tan2019efficientnet} & 0.9299 & 0.9261 & 0.9274 & 0.9261 \\
        AlexNet \cite{krizhevsky2012imagenet} & 0.9196 & 0.9174 & 0.9183 & 0.9174 \\
        DenseNet121 \cite{huang2017densely} & 0.9479 & 0.9478 & 0.9460 & 0.9478 \\
        MobileNetV2 \cite{sandler2018mobilenetv2} & \textcolor{blue}{0.9478} & \textcolor{blue}{0.9478} & \textcolor{blue}{0.9478} & \textcolor{blue}{0.9478} \\
        \hline\hline
        \multicolumn{5}{|c|}{\textbf{Parallel Hybrid CNN-Transformer with CKAN}} \\
        \hline
        % ResNet18 \cite{he2016deep} & \textcolor{blue}{0.9785} & \textcolor{blue}{0.9783} & \textcolor{blue}{0.9783} & \textcolor{blue}{0.9783} \\
        ResNet18 \cite{he2016deep} & \textcolor{blue}{\textbf{0.9785}} & \textcolor{blue}{\textbf{0.9783}} & \textcolor{blue}{\textbf{0.9783}} & \textcolor{blue}{\textbf{0.9783}} \\        
        ResNet50 \cite{he2016deep} & {0.9423} & {0.9348} & {0.9368} & {0.9348} \\
        EfficientNet-B0 \cite{tan2019efficientnet} & \textcolor{blue}{0.9452} & \textcolor{blue}{0.9435} & \textcolor{blue}{0.9441} & \textcolor{blue}{0.9435} \\
        AlexNet \cite{krizhevsky2012imagenet} & \textcolor{blue}{0.9353} & \textcolor{blue}{0.9348} & \textcolor{blue}{0.9350} & \textcolor{blue}{0.9348} \\
        DenseNet121 \cite{huang2017densely} & \textcolor{blue}{0.9698} & \textcolor{blue}{0.9696} & \textcolor{blue}{0.9697} & \textcolor{blue}{0.9696} \\
        MobileNetV2 \cite{sandler2018mobilenetv2} & 0.9470 & \textcolor{blue}{0.9478} & 0.9470 & \textcolor{blue}{0.9478} \\
        \hline
    \end{tabular}
\end{table}

\section{Conclusion}
 
In this study, we presented a hybrid CNN-Transformer approach for skin cancer classification, combining the strengths of CNNs for local feature extraction and Transformers for capturing long-range dependencies. Additionally, we introduced CKAN-based feature fusion to enhance flexibility and improve feature representation. Our experiments across multiple datasets demonstrated that hybrid models consistently outperformed CNN-only architectures, highlighting the importance of integrating spatial and contextual information for accurate medical image classification. The findings provide insights into the trade-offs between different deep learning architectures, emphasizing their relevance in handling real-world challenges such as class imbalance and dataset variability. This work contributes to the development of more reliable and generalizable AI-driven diagnostic tools for skin cancer detection.

%\section*{df}

{\small
\bibliographystyle{IEEEtran}

}

\end{document}